# Towards Trustworthy Web Attack Detection: An Uncertainty-Aware Ensemble Deep Kernel Learning Model

Yonghang Zhou, School of Management, Hefei University of Technology and Key Laboratory of Process Optimization and Intelligent Decision-making, Ministry of Education

Hongyi Zhu, Department of Information Systems and Cyber Security, University of Texas at San Antonio

Yidong Chai*, School of Management, Hefei University of Technology and Key Laboratory of Process Optimization and Intelligent Decision-making, Ministry of Education

Yuanchun Jiang, School of Management, Hefei University of Technology and Key Laboratory of Process Optimization and Intelligent Decision-making, Ministry of Education

Yezheng Liu, School of Management, Hefei University of Technology and Key Laboratory of Process Optimization and Intelligent Decision-making, Ministry of Education

Web attacks are one of the major and most persistent forms of cyber threats, which bring huge costs and losses to web application-based businesses. Various detection methods, such as signature-based, machine learning-based, and deep learning-based, have been proposed to identify web attacks. However, these methods either (1) heavily rely on accurate and complete rule design and feature engineering, which may not adapt to fast-evolving attacks, or (2) fail to estimate model uncertainty, which is essential to the trustworthiness of the prediction made by the model. In this study, we proposed an Uncertainty-aware Ensemble Deep Kernel Learning (UEDKL) model to detect web attacks from HTTP request payload data with the model uncertainty captured from the perspective of both data distribution and model parameters. The proposed UEDKL utilizes a deep kernel learning model to distinguish normal HTTP requests from different types of web attacks with model uncertainty estimated from data distribution perspective. Multiple deep kernel learning models were trained as base learners to capture the model uncertainty from model parameters perspective. An attention-based ensemble learning approach was designed to effectively integrate base learners' predictions and model uncertainty. We also proposed a new metric named High Uncertainty Ratio-F Score Curve to evaluate model uncertainty estimation. Experiments on BDCI and SRBH datasets demonstrated that the proposed UEDKL framework yields significant improvement in both web attack detection performance and uncertainty estimation quality compared to benchmark models.

CCS CONCEPTS • **Information systems**➜**Data mining;** • **Computing methodologies**➜**Uncertainty quantification;** • **Security and privacy;**

**Additional Keywords and Phrases:** Web attack detection, model uncertainty, Gaussian Process, deep kernel learning, ensemble learning

This work was partially supported by the NSFC under Grants 72171071, 72293581, 72271084, 72188101, 72101079, 72293580, 71771131, 72110107003, in part by the Excellent Fund of HFUT under Grant JZ2021HGPA0060, in part by the China Scholarship Council under Grant 202306690041.
Authors' addresses: Y. Zhou, Y. Chai (Corresponding author), Y. Jiang and Y. Liu, School of Management, Hefei University of Technology, Hefei, Anhui 230009, China; Key Laboratory of Process Optimization and Intelligent Decision-making, Ministry of Education, Hefei, Anhui 230009, China; emails: yonghangzhou@mail.hfut.edu.cn, {chaiyd, ycjiang, liuyezheng}@hfut.edu.cn; H. Zhu, Department of Information Systems and Cyber Security, University of Texas at San Antonio, 1 UTSA Circle, San Antonio, TX 78249; email: hongyi.zhu@utsa.edu.

# 1 INTRODUCTION

The development of web applications has greatly promoted economic development and social progress. In 2023, 71% of businesses were conducted with web applications all over the world [1]. Despite its huge success, web applications often have many vulnerabilities that can be exploited to carry out web attacks [2]. Web attacks exploit vulnerabilities to get unrestricted access to an organization's networks, systems, private data, and other essential assets, leading to severe consequences such as sensitive information leakage and damage to the system and bringing a huge threat to the hosted web applications [3–5]. It is reported that a successful web attack can result in an average cost of $4.45M for a company [6]. Among common web attacks, several attack types exploit web application vulnerabilities via web requests, which are easy to implement, prevalent, and highly risky [7]. For example, SQL injection and Cross-site scripting (XSS) are among the top 3 threats according to the OWASP 2021 report [8] and CWE 2023 report [9]. Despite many efforts by cybersecurity experts to protect web applications from attacks, web attack activities remain active. Furthermore, detecting and mitigating web attacks becomes increasingly challenging due to the emergence of new attacks and anti-forensics evasion techniques [4]. Hence, there is a pressing need for web attack detection capabilities to protect web applications.

Existing methods for web attack detection can be categorized into three main streams. The first stream of methods is based on signatures to identify web attacks according to security rules pre-designed by cybersecurity experts [10, 11]. Though efficient and easy to implement, signature-based methods require laborious effort from cybersecurity analysts to generate rules/signatures to address evolving and emerging attacks. The second stream uses features extracted from payloads by feature engineering to learn web attack patterns [12–14]. Traditional machine learning (ML) models (e.g., Naive Bayes, Support Vector Machines (SVMs), and Decision Tree(DT)) were then employed to learn and identify web attacks from those extracted features. While achieving high detection accuracy, the performance of feature engineering-based web attack detection hinges on the accuracy and completeness of the selected features, which also require extensive domain knowledge and laborious manual effort. The third stream uses deep learning models (e.g., Convolutional Neural Networks (CNNs), Long Short-Term Memory (LSTM), and Autoencoder) to automatically learn payload representations that capture salient attack patterns to facilitate detection [15–17] without laborious rule design or feature engineering while achieving state-of-the-art performance.

However, given that cybersecurity is a high-risk application for deep learning models, ensuring their trustworthiness in web attack detection is also critical. Past literature indicated that deep learning models tend to make uninterpretable, overconfident, and sometimes incorrect and unreliable predictions, hindering the trustworthiness of deep learning models in cybersecurity applications [18, 19]. Amongst several approaches to developing trustworthy deep learning models, measuring model uncertainty has gained the broadest interest as uncertainty estimation is one of the key parts of the decision-making process [20]. Model uncertainty covers the uncertainty caused by the limitations of the model knowledge, such as model parameters and data distribution [20, 21]. Existing uncertainty-aware deep learning models can be categorized into two main streams. The first stream measures model uncertainty from the perspective of model parameters. These models learn the posterior distribution of model parameters and calculate model uncertainty by the variance of the sampled model predictions. The sampled model predictions can be obtained by Bayesian Neural Networks (BNNs) [22], MC Dropout (McDrop) [23], or Deep Ensemble Model (DeepEnsemble) [24]. The second stream estimates model uncertainty from the perspective of data distribution. These models capture model uncertainty caused by the difference in training and test data distributions. Kernel functions were used to obtain the difference between data points across distributions, and model uncertainty was measured by the variance of the predictions modeled by the Gaussian Process with the kernel functions [25–27]. Predictions of test data that are different from training data in feature space have high model uncertainty. In web attack detection applications, not only should the model parameters be well trained to ensure the performance of detection, but the data distribution should also be taken into account as web attacks continuously evolve. Therefore, the model uncertainty of deep learning-based web attack detection models should be estimated from perspectives of both model parameters and data distribution.

In this study, we develop an uncertainty-aware ensemble deep kernel learning (UEDKL) model for web attack detection. The proposed UEDKL model estimates the model uncertainty from both model parameter and data distribution perspectives. First, deep kernel learning is selected as a classifier to distinguish different types of abnormal HTTP requests from normal HTTP requests and estimate model uncertainty from the data distribution perspective. Specifically, in the deep kernel learning model, a Transformer encoder was employed to obtain the payload representation of HTTP requests, and a Gaussian Process layer estimates the model uncertainty by comparing the distribution of data points. Then, we train multiple deep kernel learning models as base learners to estimate the uncertainty from the perspective of model parameters. Such an ensembled approach enhances the performance of web attack detection. Finally, to effectively integrate the predictions and model uncertainty from base learners, we design an attention-based ensemble learner to weight each prediction class for each base learner dynamically. The final decision was made based on the weighted average of the predictions from base learners. The model uncertainty of the UEDKL model is calculated by the total variance of predictions from base learners. The proposed model was evaluated against the baselines on BDCI and SRBH datasets. Comprehensive experiments show that the proposed UEDKL model outperformed the baselines in terms of common metrics such as overall accuracy, macro precision, macro recall, macro F1 score, weighted precision, and weighted F1 score, and also provided precise model uncertainty estimation to its prediction. The model uncertainty estimation can serve as additional information that assists security analysts in prioritizing validations on potential incorrect predictions or unseen attack samples.

The main contributions of this study are four-fold. First, we contribute to the web attack detection literature by proposing a novel model that provides accurate prediction for web attack detection and effective model uncertainty estimation for the corresponding prediction. Second, our proposed attention-based ensemble learner can dynamically adjust base classifier contributions considering model uncertainty from each base classifier, enabling effective uncertainty-aware ensemble learning, which can be generalized to applications in various domains. Third, the proposed UEDKL outperformed existing baselines (i.e., machine learning and deep learning models) in web attack detection tasks on BDCI and SRBH datasets. Fourth, we propose a High Uncertainty Ratio-F Score Curve metric to evaluate the uncertainty estimation performance. We show that our model is superior compared to the recently proposed uncertainty-aware models (i.e., McDrop, DeepEnsemble, Bayes By Backprop).

The remainder of this paper is structured as follows: Section 2 provides a review of related work. Section 3 outlines the proposed UEDKL in detail. In Section 4, we evaluate the proposed model. Finally, the concluding remarks of this study are presented in Section 5.

## 2 RELATED WORKS

In this section, we review related works on existing web attack detection methods and uncertainty-aware deep learning models to guide this study.

### 2.1 Web Attack Detection

The serious consequences and persistence of web attacks call for effective methods for detecting malicious payloads. Various web attack detection methods are proposed and then can be divided into three types: signature-based, machine learning-based, and deep learning-based methods.

The signature-based methods detect web attacks by identifying known malicious patterns from payloads. The known malicious patterns are commonly constructed by cybersecurity experts with their knowledge and experience or reported by online communities (e.g., OWASP) [28]. Signature-based methods are widely employed in web application firewalls (WAFs). For instance, ModSecurity is one of the most popular open-source web application firewall platforms with signature-based methods [29]. The platform provides a rule language called "SecRules" to define and monitor known malicious patterns. Though efficient and easy to implement, the signature-based methods rely on extensive configuration by cybersecurity experts, which requires laborious efforts. The heavy reliance on known malicious patterns also leads to poor coverage rates when new attack types are derived from well-known attacks [28].

The deficiencies of the signature-based methods motivated the machine learning-based methods where detection is based on feature engineering. Traditional machine learning models, such as Naive Bayes [14], SVM [13], Decision Tree [12], Random Forest [30], and CatBoost [31], were employed to learn and identify web attacks through the features extracted from payloads. The extracted feature includes statistic features (such as "number of keywords" and "length of payloads") [30], special character features (such as "*" and "$") [13], and bag-of-word-based features (such as TF-IDF) [12]. While achieving promising results, the performance of web attack detection is strongly influenced by the accuracy and completeness of feature extraction, which is laborious and requires domain knowledge and engineering skills. Moreover, these features are shallow and often subject to concerns regarding comprehensiveness and generalizability [32].

Due to the superior ability to learn high-dimensional complex patterns from unstructured data, deep learning models have drawn considerable attention in web attack detection tasks. Deep learning models have thrived in web attack detection. Common baseline models include Convolution Neural Network (CNN) [17], Long Short-Term Memory (LSTM) [15], and Autoencoder [16]. These models employ multiple layers of non-linear processing to learn attack patterns from payloads. Other studies combine multiple neural network structures to improve the performance of web attack detection. For instance, Niu et al. proposed a CNN-GRU model that combined CNN and Gate Recurrent Unit (GRU) to capture local features and contextual information from payloads [33]. Luo et al. proposed an ensemble deep learning model (EDL) that employs three basic neural networks, CNN, LSTM, and M-ResNet (MRN, a variant of residual network), to detect web attacks and applies Multilayer Perceptron (MLP) as ensemble learner to enhance performance [34]. The attention mechanism was applied to locate suspicious regions in payloads. Liu et al. proposed a Locate-Then-Detect (LTD) model that employs an attention-based deep neural network to detect malicious code in the payloads and a CNN to further analyze these regions and recognize web attacks [35]. Researchers have also applied large language models (LLM) to generate data representations for web attack detection [36, 37]. Nguyen et al.

leveraged Bidirectional Encoder Representations from the Transformers (BERT) model to extract representation and MLP for web attack detection [28]. Lakhani et al. employed BERT to detect SQL injection in payload [38].

Deep learning models are often trained with maximum likelihood, thus producing a point estimate but not an uncertainty value [39, 40]. Effective model uncertainty estimation is important because it allows practitioners to know the confidence level of predictions made by the model and allows a system to abstain from decisions due to low confidence [41]. Predictions made by deep learning-based models without model uncertainty estimation can be unreliable, hindering the trustworthiness of deep learning models in cybersecurity applications. So, it is critical to develop uncertainty-aware deep learning-based models for web attack detection.

**2.2 Uncertainty-Aware Deep Learning Models**

Enabling the measurement of model uncertainty is one of the essential considerations for developing trustworthy deep learning models [20, 21]. Generally, model uncertainty covers the uncertainty caused by the limitations of the model knowledge, such as model parameters and data distribution. In recent years, there has been a growing interest in estimating model uncertainty in deep learning models [20, 21, 42]. However, accurate estimation of model uncertainty is challenging as uncertainty values have no ground truth [21, 40, 43]. There are two streams of uncertainty-aware models. The first stream of uncertainty-aware models focuses on estimating model uncertainty from the model parameter perspective. The parameters of deep learning models are trained using stochastic gradient descent; thus, the parameters exhibit uncertainty [21, 24]. These models calculate model uncertainty by the variance of the sampled model predictions. The sampled model predictions can be obtained by Bayesian Neural Networks (BNNs) [22], McDrop [23], or the ensemble of deterministic neural networks [24]. For instance, The parameters of each layer in BNNs are assigned with probability distributions. The posterior distribution over the weights can be learned with training data. Once the posterior distribution over the weights has been estimated, the prediction can be obtained by Bayesian Model Averaging [21]. Blundell et al. proposed Bayes by Backprop (BayesByBackprop), in which the parameters of each layer are assigned with probability distributions [22]. The posterior distributions of the parameters were learned by variational inference and backpropagation. The prediction is calculated by averaging the prediction of models sampled from posterior distribution on the weights, and the model uncertainty can be obtained by the variance of the sampled model predictions. To achieve competitive performance and obtain high-quality uncertainty estimation, BNNs require heavy computations and appropriate prior distributions on weight, which requires huge computation cost [23]. To address these challenges, Yarin et al. proposed a dropout-based model, McDropout, where the dropout layer was interpreted as a variational Bayesian approximation [23]. McDropout keeps dropping neural units during inference and performs several forward passes. Each forward pass can sample a sub-model from neural networks. The predictions obtained from each sub-model were averaged to obtain the final predictions, and the variance of the predictions from sub-models was applied as an estimation of the model uncertainty. Since the dropout can be interpreted as an ensemble of different neural networks [44], Lakshminarayanan et al. proposed a scalable non-Bayesian solution, DeepEnsemble, that trains multiple deterministic neural networks with random parameter initialization as base learners and average predictions of base learners for the final prediction. The model uncertainty is captured by the variance of predictions over the base learners [24].

Another stream of uncertainty-aware models focuses on estimating model uncertainty from the perspective of data distribution. These models capture model uncertainty caused by the difference in training and test data distributions. The difference between data points was obtained by kernel functions, which were defined as covariance matrix in Gaussian Process (GP) [45]. GPs are Gaussian distributions interpreted as probabilistic mappings from an input data space to its output label space [46]. Model uncertainty was measured by the variance of the predictions modeled by GP [25–27]. Standard GPs face two critical challenges. First, GPs lack kernel functions that are capable of handling unstructured data (e.g., text data ). Second, GPs cannot effectively scale to large datasets due to the complex computation of the kernel matrix that encodes the covariance of all data in the dataset [47]. To address these issues, Wilson et al. proposed a Deep Kernel Learning (DKL) model [25]. The deep kernel learning model contains three components, including a deep neural network-based feature extractor to extract embedding features from unstructured inputs, a Gaussian Process layer that leverages the extract embedding features to model hidden features with model uncertainty, and a Softmax layer to map Gaussian Process-based hidden features to prediction outputs. To reduce the complex computation, a sparse Gaussian Process was adopted in the Deep Kernel Learning model following the idea that uses $M$ ($M \ll N$, $N$ is the number of samples in the train set) inducing data points to summarize the distribution of the whole training data [26]. Once the inducing data points are chosen or generated, the covariance between test samples and training data can be approximated by the covariance between test samples and inducing data points.

**2.3 Research Gaps and Questions**

To sum up, we identified two key research gaps from our literature review. First, though deep learning models achieve superior performance in web attack detection, they fail to provide model uncertainty estimation, which hinders the trustworthiness of the deployment of deep learning models in web attack detection applications. Second, in web attack detection applications, not only should the model parameters be well trained to ensure the performance of detection, but the data distribution should also be taken into account as web attacks are continuously evolving. Therefore, the model uncertainty of deep learning-based web attack detection models should be estimated from both model parameter and data distribution perspectives.

These limitations motivate the following research questions:

RQ1: How can we develop an uncertainty-aware deep learning model for web attack detection?

RQ2: How can we estimate the model uncertainty of deep learning-based web attack detection models from both model parameters and data distribution perspectives?

**3 PROPOSED METHOD**

This section begins by describing the general framework, known as uncertainty-aware ensemble deep kernel learning. We then define and formalize the details of base and ensemble learners.

**3.1 Overall Framework**

We propose an uncertainty-aware ensemble deep kernel learning (UEDKL) framework for web attack detection. The proposed UEDKL model estimates the model uncertainty from both model parameter and data distribution perspectives. First, deep kernel learning is applied as the base learner to distinguish different types of abnormal HTTP requests from normal HTTP requests and estimate model uncertainty from the data distribution perspective. Then, we train multiple deep kernel learning models as base learners to estimate the uncertainty from the perspective of model parameters and enhance the performance of web attack detection. Finally, in order to effectively integrate the predictions and model uncertainty from base models, we design an attention-based ensemble learner to weight each prediction class for each base model dynamically. The proposed UEDKL was trained in two stages: training the base classifiers and training the ensemble learner. The overall framework is depicted in Fig. 1.

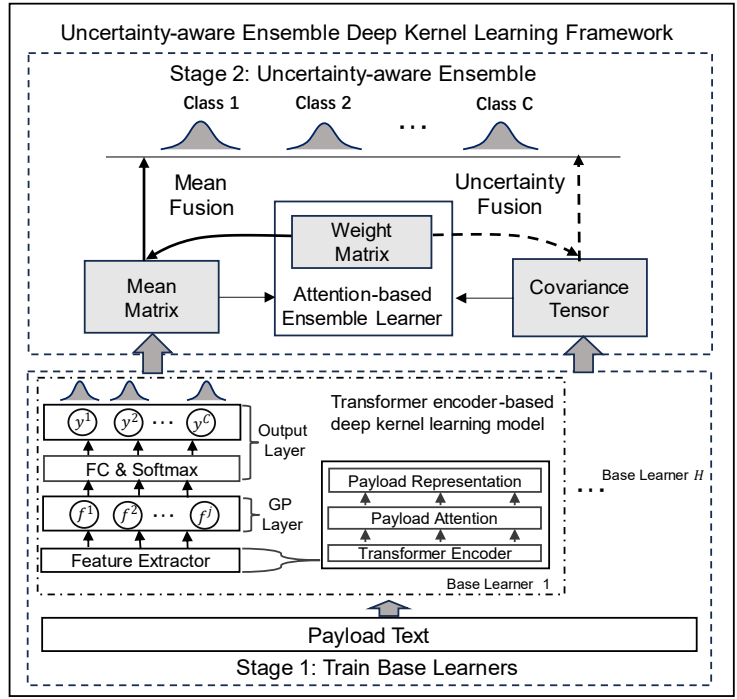

Fig. 1. The proposed UEDKL framework

### 3.2 Transformer Encoder-based Deep Kernel Learning Model

As reviewed in related work, there are three components in the deep kernel learning model: (1) a deep learning model-based feature extractor to encode deep representation from the payload data, (2) a Gaussian Process layer to represent hidden features with model uncertainty from the perspective of data distribution, and (3) a softmax layer that maps Gaussian Process hidden features to classification probability distributions.

Without loss of generality, we denote $N$ payloads and their labels as $D = \{(x, y)_i\}_{i=1}^{N}$. Each payload $x$ consists of a sequence of characters, thus can be regarded as a document containing a list of tokens $x = \{w_1, w_2, \dots, w_T\}$. As the Transformer encoder has excellent text data representation capability [48, 49], we adopted it to obtain the representati- on of payloads in the DKL model. The Transformer encoder comprises a stack of encoding units, each containing a multi-head self-attention mechanism and a fully connected feed-forward network. The encoder, which is parameterized by $\theta_{encoder}$, encodes each token $w_t$ in payload $x$ as a continuous vector $e_t$ with semantic information. Within the payload, there are both legitimate tokens and potentially harmful tokens, as the payload may contain concealed malicious code. So, each token in the payload holds varying significance in detection modeling [35]. The attention mechanism has the ability to acquire the significance of individual tokens in text modeling [50]. Therefore, the attention mechanism was employed to calculate the importance score $a_t$ of each token $e_t$.

$$r_t = \tanh(W_r e_t + b_r) \tag{1}$$

$$a_t = \frac{\exp(r_t^T w_r)}{\sum_t \exp(r_t^T w_r)} \tag{2}$$

Where $W_r, b_r, w_r$ are trainable parameters. The representation of payloads $e_x$ can be obtained by the weighted sum of all representations of its tokens with Equation (3).

$$e_x = \sum_t a_t e_t \tag{3}$$

The Gaussian Process layer comprises $J$ Gaussian Process units, $f_x = \{f_x^1, f_x^2, \dots, f_x^J\}$, representing the hidden feature with model uncertainty. $f_x^j$ is a latent function of representation $e_x$ distributed as a Gaussian Process.

$$f_x^j \sim GP(m(e_x), k(e_x, e_X)) \tag{4}$$

Where $m(e_x)$ is the mean function of $e_x$, which is often taken as zero [47], $k(e_x, e_X)$ is the kernel function that calculates the covariance between sample $x$ and all observed data $X$. The choice of an appropriate kernel requires that closer data samples are expected to be more similar than samples that are further apart. The radial basis function kernel is a widely used choice for kernel functions, denoted as Equation (5).

$$k(e_x, e_X) = \exp\left(\frac{\|e_x - e_X\|^2}{\gamma}\right) \tag{5}$$

Where $\gamma$ is the parameter that can be optimized during the training process. If data $x$ is distant from all observed data X that the model was trained on, the model has less knowledge to accurately predict the outcome for data sample $x$ compared to a scenario where $x$ is close to all observed data X. Therefore, the latent function $f_x^j$ exhibits a high variance, indicating a high level of uncertainty in the model.

In order to map the Gaussian Process's hidden features $f_x$ to the predction output, which consists of a set of mean functions and covariance functions, we need sample hidden features $\tilde{f}_x = \{f_x^{(1)}, f_x^{(2)}, \dots f_x^{(T)}\}$ for $T$ times from $f_x$, and pass $\tilde{f}_x$ to the softmax layer to get distributions of prediction $y'_x = \{y_x'^{(1)}, y_x'^{(2)}, \dots, y_x'^{(T)}\}$. The classification probability distribution can be obtained by Equation (6).

$$y_x'^{c(t)} = \frac{\exp(W_s f_x^{(t)})^c}{\sum_{i=1}^{i=C}(\exp(W_s f_x^{(t)}))^i} \tag{6}$$

Where $y_x'^{c(t)}$ denotes the prediction of $c$-th class for payload $x$ in $t$-th sampling and $W_s$ is the trainable parameters. The prediction $\hat{y}'_x$ can be calculated by the mean of $y_x'^{(t)}$ as Equation (7).

$$\hat{y}'_x = \frac{1}{T} \sum_t y_x'^{(t)} \tag{7}$$

According to prior research [51], the model uncertainty for prediction of class $c$ can be calculated by the variance of $y_x'^c$ over $T$ times sampling as Equation (8).

$$\sigma(y_x'^c) = \frac{1}{T}\sum_{t=1}^{T}(y_x'^{c(t)} - \hat{y}_x'^c)^2 \tag{8}$$

Where $\hat{y}_x'^c$ is the mean of the probability of $c$-th class prediction from $\hat{y}_x'$. Because of the operation of softmax, the probability of each class prediction is not independent. So, we calculate the covariance matrix to gain more information for the subsequent ensemble learning step by Equation (9).

$$Cov(\boldsymbol{y}_x') = \begin{pmatrix} \sigma(y_x'^1) & \cdots & cov(y_x'^1, y_x'^C) \\ \vdots & \ddots & \vdots \\ cov(y_x'^C, y_x'^1) & \cdots & \sigma(y_x'^C) \end{pmatrix} \tag{9}$$

where $cov(\boldsymbol{y}_x'^{c_i}, \boldsymbol{y}_x'^{c_j})$ is the covariance of the predictions of classes $c_i$ and $c_j$, following Equation (10).

$$cov\left(y_x'^{c_i}, y_x'^{c_j}\right) = \frac{1}{T}\sum_{t=1}^{T}\left(y_x'^{c_i(t)} - \hat{y}_x'^{c_i}\right)\left(y_x'^{c_j(t)} - \hat{y}_x'^{c_j}\right) \tag{10}$$

To avoid the huge computation cost for calculating dependence between all training data in the Gaussian Process Layer, a set of inducing points placed $\overline{D} = \{(\bar{x}_i, \boldsymbol{f}_{\bar{x}_i})\}_{i=1}^{M}$ on the grids of the payload representation space was introduced to approximate the whole dataset. $f_{\bar{x}_i}^j \in \boldsymbol{f}_{\bar{x}_i}$ is the inducing variable and assigned with a Gaussian prior $p\left(f_{\bar{x}_i}^j\right) \sim \mathcal{N}\left(0, k(e_{\bar{x}_i}, e_{\bar{x}})\right)$. Hence, $p(\boldsymbol{f}_{\bar{x}_i}) = \prod_j \mathcal{N}\left(0, k(e_{\bar{x}_i}, e_{\bar{x}})\right)$. To infer the posterior distribution of $\boldsymbol{f}_{\bar{x}_i}$, variational inference with stochastic gradient training was proposed by [25]. Specifically, the inducing variables $f_{\bar{x}_i}^j$ were specified with a posterior distribution $q\left(f_{\bar{x}_i}^j\right) \sim \mathcal{N}(o_j, s_j)$, where $\{o_j, s_j\}_{j=1}^{J}$ are variational parameters that need to be learned. Then, the trainable parameters of the base learner can be learned by optimizing the evidence lower bound (ELOB). The loss of the base learner can be devised as Equation (11).

$$\mathcal{L}_1 = \mathbb{E}_{q(\boldsymbol{f}_{\bar{x}})p(\boldsymbol{f}_x|\boldsymbol{f}_{\bar{x}})}[log p(y|\boldsymbol{f}_x)] - KL[q(\boldsymbol{f}_{\bar{x}})||p(\boldsymbol{f}_{\bar{x}})] \tag{11}$$

Where $KL$ refers to the Kullback–Leibler divergence that quantifies the distance between two distributions. All the trainable parameters of the base learner can be updated by the adaptive moment estimation (Adam) algorithm [52].

$$\theta_{baselearner} \leftarrow Adam(\nabla_{\theta_{baselearner}}\mathcal{L}_1) \tag{12}$$

Where $\theta_{baselearner} = \{\theta_{encoder}, \boldsymbol{W}_r, \boldsymbol{b}_r, \boldsymbol{w}_r, \boldsymbol{W}_s, \gamma, \{o_j, s_j\}_{j=1}^{J}\}$ denotes all the trainable parameters of the base learner.

Inspired by deep ensemble learning [24], we train $H$ transformer encoder-based deep kernel learning models with random parameter initialization for capturing the model uncertainty from the perspective of model parameters. The same training process described above is applied to all $H$ base learners. Then, we can get predictions $\boldsymbol{U} = [\boldsymbol{u}^1, \boldsymbol{u}^2, \ldots, \boldsymbol{u}^C]$, where each element $\boldsymbol{u}^c$ is the prediction of the $c$-th class from all base learners, i.e., $\boldsymbol{u}^c = [\hat{y}_1'^c, \hat{y}_2'^c, \ldots, \hat{y}_H'^c]^T$ and $\hat{y}_h'^c$ denotes the prediction of the $c$-th class from the $h$-th base learner. We can also get a covariance matrix list $\boldsymbol{\mathcal{E}} = [\boldsymbol{\Sigma}_1, \boldsymbol{\Sigma}_2, \ldots, \boldsymbol{\Sigma}_H]$, where each element $\boldsymbol{\Sigma}_h$ is the covariance matrix from the $h$-th base learner $\boldsymbol{\Sigma}_h = Cov(\boldsymbol{y}')_h$.

### 3.3 Attention-based Ensemble Learner

This section describes the details of combining the predictions of the base learners with the model uncertainty to obtain the ultimate prediction and the overall model uncertainty. We utilize the ensemble learning method [53] to integrate base learners' predictions. When combining predictions of the base learners, it is important to consider the model uncertainty. Predictions with high model uncertainty are less reliable and should be given less importance when combining them with other predictions. Conversely, predictions with low model uncertainty are more reliable and should be given greater importance. Therefore, we hope the weights of the prediction should be smaller when the prediction exhibits higher model uncertainty. To achieve this goal, we design an attention-based neural work to learn the weights of prediction of each class from each base learner, as the attention mechanism is commonly employed to learn the dynamic weights of different parts of features [35, 54]. And, we add model uncertainty of the whole model into the loss function as a regularization term to reduce the whole model uncertainty. Fig. 2 illustrates the structure of the ensemble learner.

Specifically, for each class $c$, the weight score $\boldsymbol{\alpha}^c = [\alpha_1^c, \alpha_2^c, \ldots, \alpha_H^c]^T$ of the base classifiers can be calculated by Equation (13)-(15).

$$\boldsymbol{K} = \boldsymbol{W}_k flat(\boldsymbol{\mathcal{E}}) + \boldsymbol{b}_k \tag{13}$$

$$\boldsymbol{Q}^c = \boldsymbol{W}_q \boldsymbol{u}^c + \boldsymbol{b}_q \tag{14}$$

$$\boldsymbol{\alpha}^c = softmax\left(\frac{\boldsymbol{K}\boldsymbol{Q}^{cT}}{\sqrt{d}}\right) \tag{15}$$

Where $flat$ is an operation that flattens each covariance matrix $\Sigma_h$ in $\Sigma$ to a vector. $\theta_{ensemble} = \{\boldsymbol{W}_k, \boldsymbol{b}_k, \boldsymbol{W}_q, \boldsymbol{b}_q\}$ is trainable parameters. $d$ is the dimension of $\boldsymbol{K}$. The final prediction $\hat{y}^c$ for class $c$ can be calculated by the weighted sum of $[\hat{y}_1'^c, \hat{y}_2'^c, \ldots, \hat{y}_H'^c]^T$ as Equation (16).

$$\hat{y}^c = \sum_{h=1}^H \alpha_h^c \hat{y}_h'^c \tag{16}$$

Meanwhile, we calculate the model uncertainty $\sigma(\hat{y}^c)$ for $\hat{y}^c$ by computing its variance as Equation (17).

$$\sigma(\hat{y}^c) = \sum_{h=1}^H \alpha_h^{c\,2} \sigma(y_h'^c) + 2\sum_h^H \sum_{h'}^H \alpha_h^c \alpha_{h'}^c E(y_h'^c - \hat{y}_h'^c)(y_{h'}'^c - \hat{y}_{h'}'^c) \tag{17}$$

For the loss function of the ensemble learner $\mathcal{L}_2$, three components were considered: (1) a cross-entropy term to penalize incorrect predictions, (2) a regularization term to minimize model uncertainty, and (3) an L2 regularization term to avoid overfitting. The loss function $\mathcal{L}_2$ is then denoted as Equation (18).

$$\mathcal{L}_2 = -\sum_{(x,y)\in D} \boldsymbol{y}^T \log\left(softmax(\hat{\boldsymbol{y}}_x')\right) + \delta \sum_{(x,y)\in D} \sum_{c=1}^C \sigma(\hat{y}^c) + \zeta\|\theta_{ensemble}\|_2 \tag{18}$$

Where $\delta, \zeta$ are the coefficients of the uncertainty term and L2 regularization term. Adam algorithm is adopted to optimize $\mathcal{L}_2$ as Equation (19).

$$\theta_{ensemble} \leftarrow Adam(\nabla_{\theta_{ensemble}} \mathcal{L}_2) \tag{19}$$

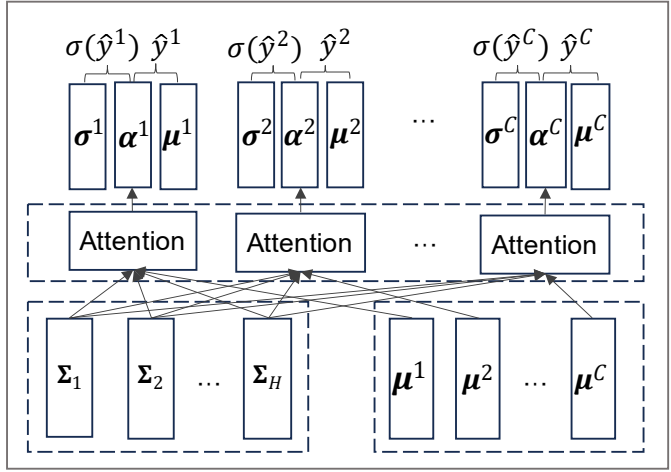

Fig. 2. The proposed attention-based ensemble learner

## 4 EVALUATION

This section begins by providing an introduction to the datasets, baseline methods, evaluation metrics, and implementation details. Then, we empirically evaluate the effectiveness of our proposed model by comparison analysis, ablation study, sensitivity analysis, and uncertainty analysis. Finally, a case study was conducted to demonstrate how our proposed model makes uncertainty-aware predictions for web attack detection tasks.

### 4.1 Research Testbed

The BDCI dataset was released by one of the largest big data competition platforms in China (https://www.datafountain.cn/). The dataset contain payloads of HTTP and was labeled with six types of attacks. To ensure that our model is capable of distinguishing normal requests and different types of attack requests, we gathered normal requests from the CSIC 2010 dataset [55], which was only labeled with normal and anomalous requests. The normal requests were added to the BDCI dataset, resulting in a total of seven type labels in our BDCI testbed. The SRBH dataset was collected during 12 days in July 2020 and was labeled with 13 type labels [31]. Similar attack patterns are grouped into seven types of attacks as in prior studies [28]. Finally, a sampled subset of the SRBH dataset with seven labels was adopted in this study. The descriptive information of the datasets is summarized in Table 1.

We applied the same preprocessing pipeline on both datasets. First, escape characters were recovered to normal characters. Second, the payload was split by punctuation. Punctuations were removed to get clean text. Finally, the clean text was split into bi-gram, trigram, and words. All characters were converted to lowercase.

### 4.2 Baselines

In order to demonstrate the efficacy of the proposed UEDKL model, we conduct comparative analysis of its performance against five conventional machine learning models, four deep learning models, and three uncertainty-aware deep learning models. The machine learning models used in this study include Naïve Bayes (NB) [56], Support Vector Machine (SVM) [13], Decision Tree (DT) [12], Random Forest (RF) [30], and Catboost [31]. These models are widely used in web attack detection and classification tasks [4, 57]. It is worth noting that RF and CatBoost are also ensemble learning models. The term frequency-inverse document frequency (TF-IDF) was applied as a feature for these machine learning models. The deep learning baselines consist of LTD [35], CNN-GRU [33], BERT [28], and EDL [34]. The uncertainty-aware models include McDrop [23], DeepEnsemble [24], and BayesByBackprop [22]. For a fair comparison, all three uncertainty-aware models employ the transformer encoder as their network architecture.

Table 1. Summary of Datasets

| BDCI | | SRBH | |
| --- | --- | --- | --- |
| Label | Count | Label | Count |
| SQL Injection | 13,991 | SQL Injection | 7,263 |
| Remote Code Execution | 9,479 | Normal | 6,191 |
| Normal | 6,000 | Fake the Source Data | 5,803 |
| Whitelist Attack | 4,831 | HTTP Response Splitting | 5,738 |
| Directory Traversal | 1,107 | Path Traversal | 5,568 |
| XSS | 689 | Code Injection | 5,130 |
| Code Execution | 657 | Manipulation | 3,074 |

### 4.3 Evaluation Metrics

Consistent with prior studies, the detection performance of models was evaluated by six metrics, including overall accuracy ($Acc$), macro precision ($P_{macro}$), macro recall ($R_{macro}$), macro F1 score ($F_{macro}$), weighted precision ($P_{weight}$), and weighted F1 score ($F_{weight}$). $F_{macro}$ and $P_{macro}$ assign the same weight to each class, while $F_{weight}$ and $P_{weight}$ consider class imbalance with the weighted average. The six metrics are defined as Equation (20)-(25).

$$Acc = \frac{TP + TN}{TP + FP + TN + FN} \quad (20)$$

$$P_{macro} = \frac{1}{C}\sum_{c=1}^{C} \frac{TP_c}{TP_c + FP_c} \quad (21)$$

$$R_{macro} = \frac{1}{C}\sum_{c=1}^{C} \frac{TP_c}{TP_c + FN_c} \quad (22)$$

$$F_{macro} = \frac{2}{C}\sum_{c=1}^{C} \frac{P_{macro}^c \times R_{macro}^c}{P_{macro}^c + R_{macro}^c} \quad (23)$$

$$P_{weight} = \frac{1}{C}\sum_{c=1}^{C} \frac{N_c}{\sum_c N_c} \frac{TP_c}{TP_c + FP_c} \quad (24)$$

$$F_{weight} = \sum_{c=1}^{C} \frac{N_c}{\sum_c N_c} \frac{P_{macro}^c \times R_{macro}^c}{P_{macro}^c + R_{macro}^c} \quad (25)$$

Where $C$ represents the number of classes, and $c$ represents the $c$-th class.

To evaluate the effectiveness of the model uncertainty, we proposed a new metric named **High Uncertainty Ratio-F Score Curve** to measure the effectiveness of the model uncertainty. Assuming that predictions with high model uncertainty (over a threshold) would be further examined (for correct predictions) and corrected (for incorrect predictions) by security analysts. We will obtain the best web attack detection performance and efficiency if incorrect predictions tend to have high uncertainty, such that analysts can review fewer payloads but correct most incorrect predictions. The High Uncertainty Ratio-F Score Curve emulates this process by recording the model performance

Algorithm 1: Calculation of High Uncertainty Ratio-F score Curve
---

**Input:** Labels of the test set: $labels = \{y_1, y_2, ..., y_N\}$
    Predictions of the test set: $preds = \{\hat{y}_1, \hat{y}_2, ..., \hat{y}_N\}$, Uncertainties of $preds$: $u = \{u_1, u_2, ..., u_N\}$, N is the size of the test set

Sort by uncertainty $u$ in descending order: $u_{sorted} = sorted(u)$

Split $u\_sorted$: $split\_points = split(u\_sorted, num=20)$

$curve\_points = \{\}$

$number\_corrected = 0$

for $p = 0, p < 20, p += 1$:
    for $i = 0, i < N, i += 1$:
        if $u_i > split\_points_p$:
            $\hat{y}_i = y_i$
            $number\_corrected += 1$
    $ratio\_of\_corrected_p = number\_corrected / N$
    $F^p_{weight} = F\_score(labels, preds, "weight")$
    $curve\_points \leftarrow (ratio\_of\_corrected_p, F^p_{weight})$

**Output:** $curve\_points$

---

gain when incorrect predictions above the uncertainty threshold are corrected. We gradually reduce the uncertainty threshold and use the the weighted F1 score $F_{weight}$ to measure the performance, the points on the curve can be calculated by algorithm 1.

### 4.4 Implementation Details

Consistent with common practice [58], we randomly split the dataset into training, validation, and testing sets following the standard ratio of 8:1:1. To evaluate the effectiveness of uncertainty-aware on unseen attack types, we reserve the data with Code Execution and Manipulation labels as the unseen types for CSIC 2010 dataset and BDCI dataset, respectively. All experiments were conducted using pytorch and scikit-learn libraries on a workstation equipped with intel xeon silver 4215R CPU and nvidia 3090 GPU.

### 4.5 Overall Performance

In this section, we report the results of the proposed UEDKL model compared to the twelve benchmarks on two datasets. Each model was run ten times, and a t-test of the mean value of the metrics was conducted to examine the statistical significance. The experimental results are presented in Tables 2 and 3. As shown in the results, our proposed UEDKL model achieved the highest performance on both datasets with trigram input. Specifically, UEDKL achieved the best performance with an accuracy of 0.9832, macro precision of 0.9809, macro recall of 0.9651, macro F1 score of 0.9726, weighted precision of 0.9834, and weighted F1 score of 0.9831 on the BDCI dataset. On the SRBH dataset, UEDKL achieved an accuracy of 0.9174, macro precision of 0.9269, macro recall of 0.9196, macro F1 score of 0.9210, weighted precision of 0.9263, and weighted F1 score of 0.9195. Therefore, the proposed UEDKL model outperformed all machine learning-based, deep learning-based models and uncertainty aware models in web attack detection tasks.

Table 2: Performance Comparison with Baseline Methods on the BDCI Dataset

|  | Models | $Acc$ | $P_{macro}$ | $R_{macro}$ | $F_{marco}$ | $P_{weight}$ | $F_{weight}$ |
|---|---|---|---|---|---|---|---|
| Classical Machine Learning-based Models | NB | 0.8535*** | 0.8046*** | 0.6872*** | 0.6848*** | 0.8886*** | 0.8525*** |
|  | SVM | 0.8704*** | 0.8141*** | 0.7586*** | 0.7545*** | 0.8959*** | 0.8713*** |
|  | DT | 0.8831*** | 0.9290*** | 0.7147*** | 0.7683*** | 0.9155*** | 0.8736*** |
|  | RF | 0.9047*** | 0.8835*** | 0.8364*** | 0.8483*** | 0.9205*** | 0.9069*** |
|  | CatBoost | 0.9149*** | 0.9271*** | 0.8062*** | 0.8377*** | 0.9223*** | 0.9118*** |
| Deep Learning-based Models | LTD | 0.9678*** | 0.9479*** | 0.9341*** | 0.9405*** | 0.9675*** | 0.9675*** |
|  | CNN-GRU | 0.9640*** | 0.9369*** | 0.9256*** | 0.9303*** | 0.9640*** | 0.9637*** |
|  | BERT | 0.9701*** | 0.9483*** | 0.9246*** | 0.9345*** | 0.9699*** | 0.9696*** |
|  | EDL | 0.9690*** | 0.9450*** | 0.9213*** | 0.9311*** | 0.9686*** | 0.9684*** |
| Uncertainty Aware Models | McDrop | 0.9668*** | 0.9347*** | 0.9360*** | 0.9351*** | 0.9670*** | 0.9668*** |
|  | DeepEnsemble | 0.9772*** | 0.9683*** | 0.9509*** | 0.9589*** | 0.9772*** | 0.9770*** |
|  | BayesByBackprop | 0.9291*** | 0.6823*** | 0.6585*** | 0.6476*** | 0.8965*** | 0.9078*** |
| UEDKL | UEDKL-bigram | 0.9827* | 0.9795*** | **0.9656** | 0.9722 | 0.9830* | 0.9827* |
|  | UEDKL-trigram | **0.9832** | **0.9809** | 0.9651 | **0.9726** | **0.9834** | **0.9831** |
|  | UEDKL-words | 0.9783*** | 0.9624*** | 0.9605** | 0.9612*** | 0.9784*** | 0.9783*** |

Note: *: p-value < 0.05, **: p-value<0.01, ***: p-value < 0.001.

Table 3: Performance Comparison with Baseline Methods on the SRBH Dataset

|  | Models | $Acc$ | $P_{macro}$ | $R_{macro}$ | $F_{marco}$ | $P_{weight}$ | $F_{weight}$ |
|---|---|---|---|---|---|---|---|
| Classical Machine Learning-based Models | NB | 0.8161*** | 0.8470*** | 0.8184*** | 0.8257*** | 0.8479*** | 0.8245*** |
|  | SVM | 0.8342*** | 0.8596*** | 0.8358*** | 0.8418*** | 0.8592*** | 0.8405*** |
|  | DT | 0.8364*** | 0.8572*** | 0.8389*** | 0.8441*** | 0.8577*** | 0.8427*** |
|  | RF | 0.8683*** | 0.9014*** | 0.8694*** | 0.8766*** | 0.8979*** | 0.8741*** |
|  | CatBoost | 0.8934*** | 0.9192*** | 0.8953*** | 0.9007*** | 0.9161*** | 0.8981*** |
| Deep Learning-based Models | LTD | 0.9076*** | 0.9135*** | 0.9076*** | 0.9092*** | 0.9135*** | 0.9092*** |
|  | CNN-GRU | 0.9002*** | 0.9066*** | 0.9002*** | 0.9019*** | 0.9066*** | 0.9019*** |
|  | BERT | 0.9133*** | 0.9219*** | 0.9156*** | 0.9167*** | 0.9219*** | 0.9154*** |
|  | EDL | 0.9123*** | 0.9200*** | 0.9149*** | 0.9156*** | 0.9202*** | 0.9143*** |
| Uncertainty Aware Models | McDrop | 0.9035*** | 0.9088*** | 0.9059*** | 0.9056*** | 0.9091*** | 0.9045*** |
|  | BayesByBackprop | 0.8589*** | 0.8861*** | 0.8580*** | 0.8581*** | 0.8805*** | 0.8559*** |
|  | DeepEnsemble | 0.9141** | 0.9225*** | 0.9166* | 0.9175*** | 0.9226*** | 0.9161** |
| UEDKL | UEDKL-bigram | 0.9155** | 0.9253*** | 0.9173* | 0.9190** | 0.9250* | 0.9178** |
|  | UEDKL-trigram | **0.9174** | **0.9269** | **0.9196** | **0.9210** | **0.9263** | **0.9195** |
|  | UEDKL-words | 0.9044*** | 0.9122*** | 0.9044*** | 0.9064*** | 0.9122*** | 0.9064*** |

Note: *: p-value < 0.05, **: p-value<0.01, ***: p-value < 0.001

Meanwhile, compared with machine learning-based models, deep learning-based models achieve better detection performance. This implies the advantages of deep representations compared to machine learning-based models.

### 4.6 Ablation Study

An ablation study was conducted to validate the contribution of each proposed module in UEDKL. Concretely, we separately removed (1) the attention module of payload representation, (2) the ensemble module, (3) the uncertainty input for the ensemble module, and (4) the covariance between each label prediction in the covariance matrix $\Sigma_h$

Table 4: Performance Comparison with UEKDL's Variants

| Datasets | Variants | $Acc$ | $P_{macro}$ | $R_{macro}$ | $F_{marco}$ | $P_{weight}$ | $F_{weight}$ |
|---|---|---|---|---|---|---|---|
| BDCI | w/o_attention | 0.9821** | 0.9768*** | 0.9615*** | 0.9686*** | 0.9823** | 0.9820** |
|  | w/o_ensemble | 0.9648*** | 0.9455*** | 0.9306*** | 0.9373*** | 0.9647*** | 0.9646*** |
|  | w/o_cov | 0.9829 | 0.9775** | 0.9627** | 0.9696*** | 0.9827** | 0.9824** |
|  | w/o_uncertainty | 0.9823** | 0.9769*** | 0.9621*** | 0.9690*** | 0.9825** | 0.9822** |
|  | UEDKL | **0.9832** | **0.9809** | **0.9651** | **0.9726** | **0.9834** | **0.9831** |
| SRBH | w/o_attention | 0.9150** | 0.9214*** | 0.9178 | 0.9179*** | 0.9219*** | 0.9166** |
|  | w/o_ensemble | 0.9019*** | 0.9076*** | 0.9019*** | 0.9031*** | 0.9076*** | 0.9031*** |
|  | w/o_cov | 0.9153** | 0.9235*** | 0.9186 | 0.9187** | 0.9235*** | 0.9173** |
|  | w/o_uncertainty | 0.9140** | 0.9221*** | 0.9165* | 0.9173*** | 0.9222*** | 0.9160** |
|  | UEDKL | **0.9174** | **0.9269** | **0.9196** | **0.9210** | **0.9263** | **0.9195** |

Note: *: p-value < 0.05, **: p-value<0.01, ***: p-value < 0.001.

from the proposed model. We denoted the four variants as w/o_attention, w/o_ensemble, w/o_uncertainty, and w/o_cov, respectively. We evaluated the performance of each variant on accuracy, macro precision, macro recall, macro F1 score, weighted precision, and weighted F1 score. The results of the ablation study are shown in Table 4. According to the data in Table 4, it is evident that UEDKL significantly outperforms its variants on all metrics. Specifically, with the help of the attention mechanism, the significance of each token can be calculated to obtain the representation of the payload, which improves the overall performance of web attack detection tasks. The ensemble module plays a most significant role in improving the performance of web attack detection as the designed ensemble learner can effectively integrate the prediction and model uncertainty. When it comes to the input of the ensemble module, the results indicate that both standard deviation and covariance between each label prediction in the covariance matrix $\Sigma_h$ contributed to the performance improvement. These results demonstrate the efficacy of all designed modules.

### 4.7 Sensitivity Analysis

Sensitivity analysis was carried out to analyze how the model's performance is affected by hyperparameters. These hyperparameters include the number of encoder layers, the number of encoder attention heads, the number of Gaussian Process units in the GP layer, the grid size that determines the scale of reducing points in kernel approximation, the number of Gaussian Process units in the GP layer, the grid size that determines the scale of reducing points in kernel approximation, the number of base classifiers denoted as ensemble size, and the coefficient of the uncertainty term $\delta$. The results are shown in Tables 5 and 6. The results indicate that the proposed model performed best with 2 encoder layers, 4 encoder attention heads, 512 Gaussian Process units, a grid size of 256, 6 base classifiers, and a $\delta$ value of 0.001 on BDCI datasets. The proposed model achieves best performance on the SRBH dataset while using 3 encoder layers, 4 encoder attention heads, 512 Gaussian Process units, a grid size of 384, 6 base classifiers, and a $\delta$ value of 0.001. Meanwhile, we find that the number of base learners, the number of Gaussian Process units in the GP layer, and the grid size have a greater influence on web attack detection performance compared to other hyperparameters.

### 4.8 Uncertainty Analysis

In this section, we analyze the model uncertainty estimated by our proposed UEDKL model, McDrop, DeepEnsemble,

Table 5: Sensitivity Analysis Results on the SRBH Dataset

| Parameters | Acc | $P_{macro}$ | $R_{macro}$ | $F_{marco}$ | $P_{weight}$ | $F_{weight}$ |
| --- | --- | --- | --- | --- | --- | --- |
| Encoder_Layer_1 | 0.9827 | 0.9796 | 0.9645 | 0.9716 | 0.9829 | 0.9827 |
| Encoder_Layer_2 | **0.9832** | **0.9809** | 0.9651 | **0.9726** | 0.9834 | **0.9831** |
| Encoder_Layer_3 | 0.9830 | 0.9800 | **0.9654** | 0.9724 | 0.9832 | 0.9829 |
| Encoder_heads_2 | 0.9820 | 0.9753 | 0.9606 | 0.9673 | 0.9822 | 0.9819 |
| Encoder_heads_4 | 0.9832 | **0.9809** | **0.9651** | **0.9726** | 0.9834 | **0.9831** |
| Encoder_heads_6 | **0.9833** | 0.9781 | 0.9622 | 0.9696 | **0.9835** | **0.9831** |
| GPLayer_256 | 0.9781 | 0.9651 | 0.9559 | 0.9599 | 0.9781 | 0.9780 |
| GPLayer_512 | **0.9832** | **0.9809** | **0.9651** | **0.9726** | **0.9834** | **0.9831** |
| GPLayer_768 | 0.9815 | 0.9740 | 0.9624 | 0.9678 | 0.9817 | 0.9815 |
| grid_size_128 | 0.9787 | 0.9651 | 0.9614 | 0.9629 | 0.9788 | 0.9787 |
| grid_size_256 | **0.9832** | **0.9809** | 0.9651 | **0.9726** | **0.9834** | **0.9831** |
| grid_size_384 | 0.9828 | 0.9798 | **0.9655** | 0.9723 | 0.9830 | 0.9828 |
| Ensemble_size_4 | 0.9756 | 0.9567 | 0.9507 | 0.9531 | 0.9757 | 0.9755 |
| Ensemble_size_6 | **0.9832** | **0.9809** | 0.9651 | **0.9726** | **0.9834** | **0.9831** |
| Ensemble_size_8 | 0.9826 | 0.9795 | **0.9652** | 0.9720 | 0.9829 | 0.9826 |
| $\delta$_0.005 | 0.9813 | 0.9734 | 0.9623 | 0.9674 | 0.9814 | 0.9812 |
| $\delta$_0.001 | **0.9832** | **0.9809** | **0.9651** | **0.9726** | **0.9834** | **0.9831** |
| $\delta$_0.0005 | 0.9824 | 0.9777 | 0.9571 | 0.9664 | 0.9825 | 0.9822 |

Table 6: Sensitivity Analysis Results on the BDCI Dataset

| Parameters | Acc | $P_{macro}$ | $R_{macro}$ | $F_{macro}$ | $P_{weight}$ | $F_{weight}$ |
| --- | --- | --- | --- | --- | --- | --- |
| Encoder_Layer_2 | 0.9165 | 0.9230 | 0.9195 | 0.9196 | 0.9233 | 0.9180 |
| Encoder_Layer_3 | **0.9174** | **0.9269** | 0.9196 | **0.9210** | **0.9263** | **0.9195** |
| Encoder_Layer_4 | 0.9169 | 0.9268 | **0.9201** | 0.9205 | 0.9263 | 0.9192 |
| Encoder_heads_2 | 0.9143 | 0.9217 | 0.9143 | 0.9162 | 0.9217 | 0.9162 |
| Encoder_heads_4 | **0.9174** | **0.9269** | **0.9196** | **0.9210** | **0.9263** | **0.9195** |
| Encoder_heads_6 | 0.9161 | 0.9264 | 0.9176 | 0.9196 | 0.9260 | 0.9185 |
| GPLayer_256 | 0.9117 | 0.9190 | 0.9143 | 0.9150 | 0.9191 | 0.9136 |
| GPLayer_512 | **0.9174** | **0.9269** | **0.9196** | **0.9210** | **0.9263** | **0.9195** |
| GPLayer_768 | 0.9147 | 0.9247 | 0.9167 | 0.9184 | 0.9242 | 0.9170 |
| grid_size_256 | 0.9145 | 0.9247 | 0.9163 | 0.9182 | 0.9242 | 0.9169 |
| grid_size_384 | **0.9174** | **0.9269** | **0.9196** | **0.9210** | **0.9263** | **0.9195** |
| grid_size_512 | 0.9168 | **0.9270** | 0.9188 | 0.9204 | 0.9262 | 0.9191 |
| Ensemble_size_4 | 0.9106 | 0.9170 | 0.9137 | 0.9137 | 0.9173 | 0.9122 |
| Ensemble_size_6 | **0.9174** | **0.9269** | 0.9196 | **0.9210** | **0.9263** | **0.9195** |
| Ensemble_size_8 | 0.9171 | 0.9236 | **0.9202** | 0.9201 | 0.9239 | 0.9186 |
| $\delta$_0.005 | 0.9128 | 0.9211 | 0.9152 | 0.9162 | 0.9212 | 0.9148 |
| $\delta$_0.001 | 0.9174 | 0.9269 | 0.9196 | 0.9210 | 0.9263 | 0.9195 |
| $\delta$_0.0005 | 0.9145 | 0.9221 | 0.9170 | 0.9177 | 0.9223 | 0.9164 |

and BayesByBackprop. Fig. 3 illustrates the distribution of model uncertainty estimated by four models on correct prediction samples, incorrect prediction samples, and unseen web attack samples. We found that our proposed model exhibited a greater disparity in the distribution of model uncertainty among the three categories of samples compared to the other three models. Concretely, the model uncertainty of correct predictions estimated by the UEDKL model is significantly lower than that of incorrect prediction samples and unseen attack samples. Furthermore, the distribution

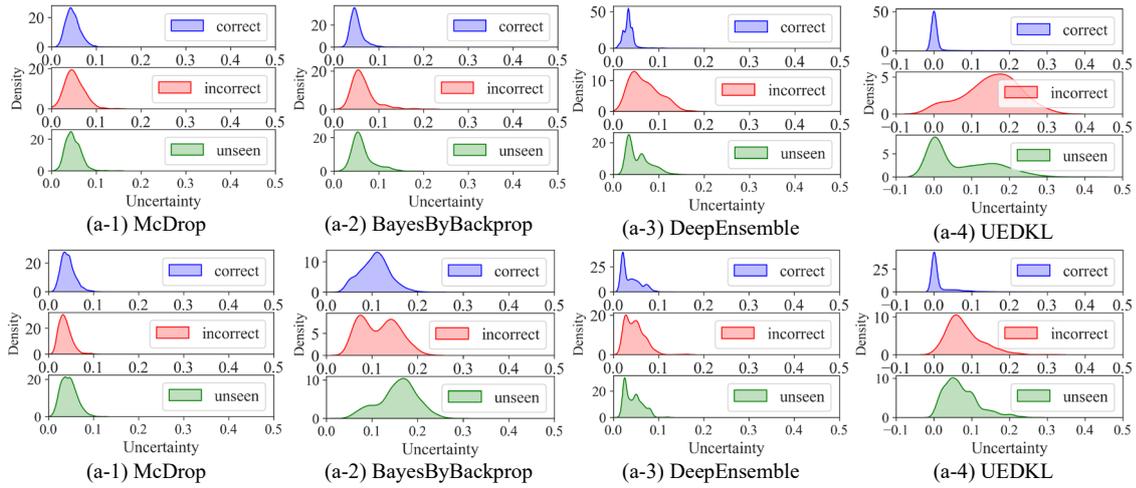

Fig. 3. The model uncertainty distribution in correctly predicted sample set incorrectly predicted sample set and unseen web attack type sample set estimated by McDrop, BayesByBackprop, DeepEnsemble, and our proposed UEDKL model on the BDCI dataset (a-1 to a-4) and the SRBH dataset (b-1 to b-4).

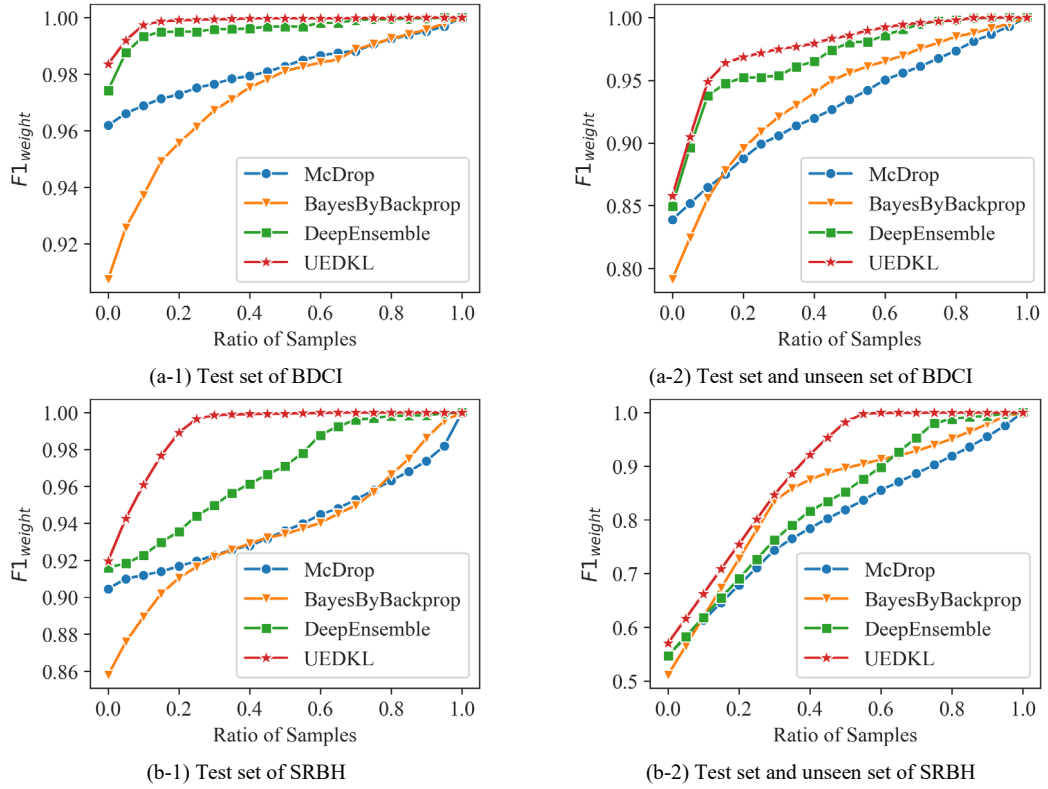

Fig. 4. The high Uncertainty Ratio-F score Curve of the test set and the mixed set (test set + unseen set) on BDCI (a-1 to a-2) and SRBH (b-1 to b-2) Datasets

of the model uncertainty for correct predictions is more concentrated around a central value than that of incorrect prediction samples and unseen type of attack samples. There are considerable overlaps between the model uncertainty distribution on three types of samples estimated by McDrop, BayesByBackprop, and DeepEnsemble. These results imply that the model uncertainty estimated by our proposed model is more precise in distinguishing correct predictions and incorrect predictions or unseen attack samples.

The same conclusion can be found in our proposed High Uncertainty Ratio-F Score curve. As the proposed UEDKL model estimates the model uncertainty from the perspective of both parameter and data distributions, Fig. 4 shows the performance of the weight F1 score of our proposed UEDKL improves much faster when more predictions with high uncertainty are manually handled (validated or corrected) in the test set compared with McDrop, BayesByBackprop, and DeepEnsemble. The weighted F1 score of our proposed UEDKL model reached 1, with the lowest ratio of high uncertainty are manually handled (validated or corrected) in the test set compared with McDrop, BayesByBackprop, and DeepEnsemble. We tested the High Uncertainty Ratio-F Score curve in the dataset that mixes the test set and unseen new attack samples; our proposed model also performed better than other models. The proposed UEDKL model captures higher model uncertainty for incorrect predictions and new attack samples than for correct predictions. When a high model uncertainty threshold is set, a small ratio of predictions needs to be further handled in which most of the predictions are wrong. Therefore, the model uncertainty estimated by our proposed model is more precise than McDrop, BayesByBackprop, and DeepEnsemble in the web attack detection task.

**4.9 Case Study**

In this section, we demonstrate how the proposed model can provide uncertainty-aware predictions in web attack detection tasks. We applied the proposed UEDKL to two request samples selected from the BDCI and SRBH test sets. As shown in Table 7, sample A, taken from the BDCI dataset, was labeled with SQL Injection, and sample B, taken from the SRBH dataset, was labeled with Code Injection. Fig. 5 illustrates the predictions of the base learners for web attack detection with model uncertainty and the weights of each label learned from the ensemble learner.

For sample A, Fig. 5(a-1) shows that all six base learners successfully classify the sample to SQL Injection (Label 1). Base learners 1, 3, and 6 assigned probability 0.75, 0.71, and 0.73 to the label of SQL Injection, respectively, which is higher than the other three base learners. Fig. 5(a-2) shows that the model uncertainty of prediction for SQL Injection estimated by base learners 1 and 6 are $0.2480 \times 10^{-2}$ and $0.2860 \times 10^{-2}$, which is significantly lower than the other four base learners. It means that base learners 1 and 6 exhibit a higher level of confidence in classifying the sample as SQL Injection. Considering both prediction probabilities and the model uncertainty, it is evident from Fig. 5(a-3) that the ensemble learner assigned a weighted score of 0.19 to the prediction of Label 1 for base learners 1 and 6, which is higher than those of the other four base learners. Finally, by averaging the prediction by weights, the prediction of the SQL Injection label was assigned the highest value, as shown in Fig. 5(a-4). From the descriptions above, our proposed UEDKL can assign high weights to the predictions with lower model uncertainty.

For sample B, we can see from Fig. 5(b-1) that base learners 1, 3, 4, 5, and 6 successfully classified the sample as Code Injection (Label 5), while base learner 2 wrongly classified the sample to SQL Injection (Label 2). Fig. 5(b-2) shows the model uncertainty of predictions for Label 5 of base learners 3, 4, 5, and 6 is 0.008, 0.012, 0.009, and 0.006, respectively, which are lower than base learners 1 and 2. This implies that base learners 3, 4, 5, and 6 are more confident in classifying the sample to Code Injection than base learners 1 and 2. Especially, while base learner 2 assigned the highest probability to SQL Injection (Label 2), the prediction has the highest model uncertainty of 0.04, which implies that classifying the example as the SQL Injection (Label 2) is less confident. It can be found in Fig.

Table 7: Examples from BDCI and SRBH Datasets

| Sample | HTTP Request Payload | Label |
|---|---|---|
| A | /api/datasources/proxy/2/query?db=telegraf&q=select mean ('total') from 'processes' where ('host' =~ /^srv75.se.zzzc.qihoo.net$/) and time >= 1658716661882ms group by time (500ms) fill (none); select mean ('zombies') from 'processes' where ('host' =~ /^srv75\.se\.zzzc\.qihoo\.net$/) and time>=1658716661882ms group by time(500ms) fill (none); | Label 1 (SQL Injection) |
| B | /blog/index.php/my-account/lost-password/user_login=euphoricleopard6&ur_reset_password=true&_wpnonce=${@print(chr(122).chr(97).chr(112).chr(95).chr(116).chr(111).chr(107).chr(101).chr(110))}&_wp_http_referer=/blog/index.php/my-account/lost-password/ | Label 5 (Code Injection) |

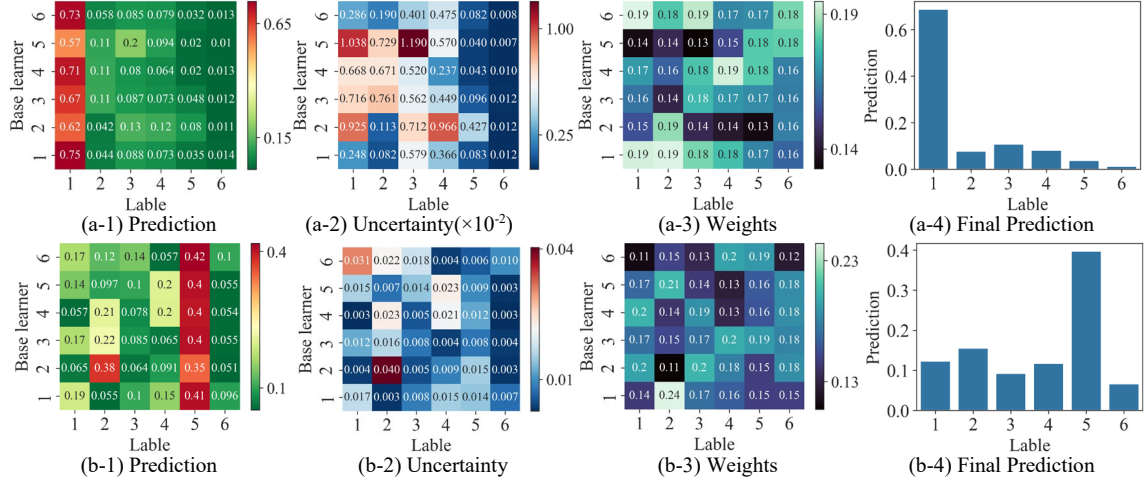

Fig. 5. Base learner's predictions, uncertainty, ensemble weights for each class prediction of base learner, and the final UEDKL predictions.

5(b-3) that the ensemble learner assigned a weighted score of 0.19 to the prediction of Label 5 for base learners 3 and 6. The wrong prediction of Label 2 for base learner 2 was assigned with a weighted score of 0.11, which reduced the influence of the wrong prediction from base learner 2 on the final prediction shown in Fig. 5(b-4). From the comparison above, we can see the Transformer encoder-based deep kernel learning model exhibits a high model uncertainty when making incorrect predictions. On the other hand, the ensemble learner is capable of assigning a lower weight score to these highly uncertain and incorrect predictions. This effectively minimizes the impact of incorrect predictions made by the base learners on the final prediction.

From the above results, it is evident that our proposed UEDKL model can effectively make predictions for web attack detection tasks with model uncertainty. The base learners tend to make correct predictions with lower model uncertainty because the Gaussian Process layer tends to generate hidden features with low variance when the sample is similar to the samples in the training set. With the help of the attention mechanism, the designed attention-based ensemble learner can effectively leverage the predictions and model uncertainty of base learners to make a final prediction.

## 5 CONCLUSION

In this paper, we address the problem of model uncertainty estimation in web attack detection tasks. Though deep learning models achieve superior performance among the existing web detection methods, the lack of model uncertainty estimation hinders the trustworthiness of deep learning models in cybersecurity applications. Therefore, we proposed an uncertainty-aware ensemble deep kernel learning model (UEDKL) to detect web attacks from HTTP request payload data with the model uncertainty captured from the perspective of both model parameters and data distribution. In addition, we propose a new metric to evaluate the estimation of model uncertainty. Experimental results on BDCI and SRBH datasets demonstrate that our proposed model significantly outperforms the baselines and provides more precise model uncertainty estimation. The model uncertainty can served as additional information for security analysts to prioritize validations on potentially incorrect predictions or unseen attack samples.

There are several promising directions for future research. First, the proposed model detects web attacks from independent HTTP request payloads; researchers can consider spatial-temporal characteristics in the feature study. Second, the proposed UEDKL model requires two-stage training, which is a time-consuming process. Hence, an end-to-end training strategy can be studied in the future. Third, this study focuses on improving the trustworthiness of deep learning-based web attack detection models from the perspective of model uncertainty estimation. There are other considerations to develop trustworthy web attack detection techniques, such as interpretability and adversarial robustness. Hence, future work can focus on developing trustworthy deep learning-based web attack detection models from the perspective of model interpretability and adversarial robustness. Finally, future work can examine the generalizability of our proposed UEDKL model to other high-risk domains, such as finance applications and health applications.